\documentclass[preprint,12pt]{elsarticle}

\usepackage{amssymb}

\usepackage{amsthm}

\usepackage{float}
\usepackage{placeins} 
\usepackage{graphicx}
\usepackage{booktabs}
\usepackage[pdfborder={0 0 0}]{hyperref}  
\usepackage{array}  
\usepackage{multirow}
\journal{Nuclear Physics B}
\usepackage{geometry}  
\usepackage{xcolor}
\begin{document}

\begin{frontmatter}



\title{HADRec: A Hierarchy-Aware Drug Recommendation Framework by Fusing Molecular Knowledge and Electronic Health Record}


\author{Junke Wang\fnref{co-first}} 
\author{Hongshun Ling\fnref{co-first}}
\author{Li Zhang\corref{cor}}
\ead{zhangli1996@163.com}
\author{Jinjing Wu}
\author{Tong Shao}
\author{Fang Wang}
\author{Yuan Gao}

\affiliation{organization={School of Biomedical Engineering,South-Central Minzu University},
            addressline={182 Minzu Avenue, Hongshan District}, 
            city={WuHan},
            postcode={430074}, 
            state={HuBei},
            country={China}}

\fntext[co-first]{Equal Contribution} 
\cortext[cor]{Corresponding author}
\begin{abstract}
Accurate medication recommendation is central to clinical decision-making, directly determining therapeutic efficacy and patient safety. However, existing methods suffer from two key limitations: drugs are often abstracted as discrete tokens, ignoring their molecular structures and pharmacological mechanisms, and the commonly used “flat” recommendation paradigm fails to leverage the hierarchical logic of the internationally standardized Anatomical Therapeutic Chemical (ATC) classification system. To address these issues, we propose HADRec, a Hierarchy-Aware Drug Recommendation framework that integrates molecular knowledge with electronic health records (EHRs). HADRec employs LLaMA-7B to encode clinical notes for rich patient representations and ChemBERTa to encode drug Simplified Molecular Input Line Entry System strings, building a global molecular knowledge base. A cross-attention mechanism then performs deep multimodal fusion between patient states and drug features. The framework further incorporates a hierarchical predictor and a novel consistency constraint loss to enforce strict adherence to ATC logical dependencies. Extensive experiments on MIMIC-III demonstrate that HADRec achieves state-of-the-art performance across Jaccard, F1, and PR-AUC. External validation on MIMIC-IV confirms strong generalization under distribution shifts, and calibration analysis shows well-calibrated predictive confidence on MIMIC-IV with ECE = 0.04, and Brier = 0.06. Counterfactual evaluation reveals clinically aligned reasoning, disentangling disease-specific treatments from general care. Together, these results establish HADRec as a high-performance, interpretable, and clinically grounded pathway toward safe and reliable AI-driven medication recommendation.
\end{abstract}

\begin{highlights}

\item First drug recommendation framework integrating molecular structures and ATC hierarchy.
\item Enables interpretable cross-modal alignment between clinical context and pharmacology.
\item Ensures prescriptions comply with medical taxonomy via hierarchical consistency constraints.
\item Achieves superior accuracy and safety in multi-label recommendation on MIMIC-III.
\item Offers a transferable, ontology-aware paradigm for knowledge-infused clinical AI.
\end{highlights}


\end{frontmatter}



\section{Introduction}
\label{sec1}

Clinical medication recommendation is a complex yet essential decision-making process that bridges medical knowledge with patient characteristics\cite{ali2023deep,yang2021safedrug}. In practice, clinicians must balance multifaceted clinical presentations, individualized physiological and pathological factors, potential drug–drug interactions (DDIs), and evolving biomedical evidence to formulate optimal treatment plans\cite{shang2019gamenet,zhou2025multi}. This high-dimensional reasoning imposes substantial cognitive demands, and prescribing risk escalate notably for patients with multiple comorbidities\cite{shang2019gamenet,li2023patient}. Against this backdrop, AI-driven medication decision support systems have emerged as a promising strategy to leverage large-scale electronic health records (EHR) data, learning precise prescribing patterns\cite{su2020gate}, reducing clinician workload, enhancing therapeutic standardization, and advancing personalized care.

Recent years have witnessed rapid progress in deep learning-based medication recommendation models that utilize EHR data, evolving from rule-based systems to sophisticated longitudinal sequence modeling\cite{dasgupta2024big,huang2024interpretable,liu2025edrmm}. These methods typically analyze patient histories, such as diagnoses, procedures and medication sequences, to infer underlying health state and predict suitable drugs. However, they are constrained by two major limitations that critically restrict their clinical reliability and generalizability:

First, they rely on abstract drug representations that disregard pharmacological mechanisms. Most existing models treat medications as discreate categorical tokens, ignoring their chemically grounded molecular structures\cite{li2025towards,WOS:001425157900001}. As a consequence, these models cannot reason about mechanistic drug actions or distinguish between therapeutically similar yet chemically distinct agents. Their predictions thus remain confined to shallow statistical associations, compromising interpretability and limiting generalizability to newly introduced drugs\cite{liu2024safe} .

Second, they adopt a flat prediction paradigm that ignores medical taxonomy.The Anatomical Therapeutic Chemical (ATC) classification system offers a rigorous, multi-level hierarchy that encodes therapeutic relevance and chemical relationships among drugs. However, prevailing models still formulate medication recommendation as a flat multi-label classification task\cite{feng2024mrndr,nguyen2025egnet}, overlooking the structural constraints embedded in the ATC system. This often leads to logically inconsistent predictions, for instance, suggesting pharmacologically incompatible drugs within higher-level therapeutic categories, which undermines clinical credibility and practical utility.

To address these challenges, this paper proposes HADRec, a Hierarchy-Aware Drug Recommendation framework. The work makes three core contributions: 

\begin{enumerate}
  \item
A novel multimodal fusion framework. We integrate LLaMA-7B–based clinical texts representations with ChemBERTa-derived molecular embeddings, achieving cross-modal alignment between patient states and pharmacological knowledge. This enables the model to ground recommendations in both clinical context and drug mechanism.
  \item
A hierarchy-consistent prediction mechanism. We develop a hierarchical prediction architecture that mirrors clinical reasoning and introduce a consistency-constrained loss function. This ensures that all recommendations adhere to the structural and logical constraints of the ATC taxonomy, enhancing clinical plausibility. 
  \item
Comprehensive empirical validation. Extensive experiments on real-word EHR datasets demonstrate the effectiveness of HADRec. Ablation studies together with case analyses further elucidate the contribution of each module and illustrate the interpretability of the framework.
\end{enumerate}

Through these contributions, HADRec provides a robust pathway toward more reliable, interpretable, and clinically grounded decision-support systems.

\section{Related Work}
\label{sec2}
Driven by the rapid advancement of predictive machine learning frameworks in healthcare, automated clinical decision-making has evolved into a sophisticated field for therapeutic discovery and disease management\cite{ahmad2024extensive}. Within this context, medication recommendation is a cornerstone of intelligent clinical decision support, aiming to generate safe, coherent, and interpretable prescriptions by synthesizing patient clinical records with pharmacological knowledge. Early studies primarily focused on modeling EHR sequences, employing recurrent neural networks\cite{liu2024safe,zhu2024rasnet}, attention-based architectures\cite{wang2025beyond}, and Transformer-style temporal models\cite{ge2025personalized} to capture longitudinal health trajectories and comorbidity patterns (e.g., SafeMed\cite{yu2024aka}, DKGM\cite{zhao2025dkgm}). Although effective at learning temporal dependencies and common prescribing patterns, these approaches typically represent medications as discrete identifiers or low-dimensional embeddings\cite{wang2025graph,wu2024promise}, overlooking their underlying chemical structures and mechanistic properties. As a result, their interpretability is limited, and their capacity to generalize to drugs unseen during training remains constrained.

To overcome this representational gap, subsequent works have integrated molecular-level information to enrich drug embeddings. The transition from static descriptors to deep learning-based molecular annotations and interpretative representations represents a pivotal shift in computational pharmacology. Advanced representation learning frameworks now enable the extraction of nuanced molecular semantics directly from structural data. Models such as SafeDrug\cite{yang2021safedrug}, DrugDoctor\cite{kuang2024drugdoctor}, and KGR\_GCLR\cite{an2024disease} utilize graph neural networks to encode molecular structures or knowledge graph relations. Other methods, including CIDGMed\cite{liang2025cidgmed} and DNMDR\cite{liu2025dnmdr}, adopt dual-granularity fusion strategies, learning representations from both molecular graphs and visit-level sequences. DEPOT\cite{zhao2024enhancing} further decomposes molecules into semantic motif trees and employs a graph transformer to model cooperative interactions among motifs. More broadly, advanced representation learning frameworks have also focused on the annotation and integration of molecular features to derive semantically enriched representations, which have been shown to facilitate improved clinical understanding and support downstream tasks such as diagnosis and therapeutic decision-making\cite{ahmad2025deep}. Despite these advances, most existing approaches treat molecular features as static priors and integrate them with patient representations through simple concatenation or similarity-based interactions\cite{li2024significance}. This shallow interaction fails to achieve deep semantic coupling between clinical context and chemical knowledge, resulting in a persistent “modality gap”: although molecular structures are incorporated, they are not dynamically aligned with evolving patient states during inference, which limits true mechanistic reasoning.

In parallel, a distinct line of inquiry has explored the integration of hierarchical information and multi-stage information. By employing comprehensive consensus modeling and systematic validation, such frameworks demonstrate how structured reasoning can maintain logical consistency across multi-level predictions\cite{yadav2023predictive}. Despite these advances, however, current medication recommendation frameworks still predominantly adopt flat multi-label formulations that overlook the intrinsic hierarchical logic of the ATC taxonomy. For instance, ChemR uses multi-task learning to jointly predict multiple ATC levels\cite{sun2023cehmr}, while SHAPE encodes hierarchical graph structures through graph neural networks\cite{chairat2024enhancing}. Nevertheless, these methods generally treat hierarchical signals as auxiliary information rather than as strict structural constraints. Their loss functions optimize each ATC level independently, without explicitly enforcing the dependency that lower-level predictions must be logically consistent with higher-level assignments\cite{deng2024hyper}. Consequently, such models may generate clinically incoherent recommendations, for instance, suggesting drugs that belong to contradictory therapeutic categories, which undermines both interpretability and real-word clinical validity. Optimizing such complex, multi-objective loss functions remains a significant challenge in therapeutic discovery, often requiring sophisticated algorithmic frameworks to balance predictive accuracy with structural consistency\cite{ahmad2025machine}.

Motivated by these observations, this work addresses two interconnected limitations in the current literature: the insufficient deep interaction between clinical and molecular representations, and the absence of rigorously enforced hierarchical constraints derived from the ATC system. To this end, we propose a hierarchy-aware drug recommendation framework that deeply fuses molecular knowledge with EHRs. The framework employs a cross-modal attention mechanism to dynamically align chemical semantics from drug Simplified Molecular Input Line Entry System (SMILES) strings with patient-specific clinical contexts. Furthermore, it incorporates a dedicated hierarchy-aware prediction architecture together with a novel consistency loss. Together, these components are designed to collectively ensure that the model not only performs enhanced pharmacological reasoning but also delivers recommendations that are semantically coherent and strictly consistent with the ATC hierarchy.
\section{HADRec Framework}
\label{sec3}

Clinical medication recommendation aims to generate safe and effective personalized prescriptions based on a patient’s EHR. In this work, we formulate the task as a multi-label classification problem augmented with ATC hierarchy constrains. Below, we formally define the key components and the core task.

\subsection{Problem Formulation}
\label{subsec1}

\subsubsection{Electronic Health Records (EHR)}
\label{subsubsec1}
An EHR dataset contains the longitudinal medical records of \(N\) patients, denoted as \(\mathcal{D}=\{P^{(1)},P^{(2)},\ldots,P^{(N)}\}\), where the superscript identifies individual patients. For a patient \(\mathcal{P}^{(i)}\), the medical history is a temporal sequence of clinical visits \(\mathcal{P}^{(i)}=\{v_{1}^{(i)},v_{2}^{(i)},\ldots,v_{T}^{(i)}\}\). Here, \(T\) is the patient’s total visit count, and each visit \(v_{t}^{(i)}\) includes a set of clinical events recorded at that visit, primarily diagnosis codes \(\mathcal{D}_{t}^{(i)}\), procedure codes \(\mathcal{P}_{t}^{(i)}\) and medications name \(\mathcal{M}_{t}^{(i)}\) .

To construct predictive inputs, we define the historical context for the $T$-th visit as the set of all preceding visits: $\mathcal{H}_T^{(i)} = \{ v_k^{(i)} \mid T - 3 \leq k < T \}$ (for patients with fewer than three prior visits, all available historical data are used),as justified in Section \ref{subsec17}.The input features for prediction consist of the complete clinical events from the $\mathcal{H}_T^{(i)}$, complemented by the diagnoses and procedures of the current visit: $\tilde{v}_T^{(i)} = \{ D_t^{(i)}, P_t^{(i)} \}$. The target output is the set of medications prescribed at the current visit, represented not by raw medication names but by their hierarchical ATC codes, as detailed in \ref{subsubsec2}.
\subsubsection{Medications and the ATC Hierarchy}
\label{subsubsec2}
Let \( M = \{m_1, m_2, \ldots, m_{|M|}\} \) denote the set of all unique medications in the dataset, where \( |M| \) is the total drug count. Each medication \( m_j \) is associated with an ATC code, which follows a strict K-level hierarchical taxonomy (we use \( K = 4 \) ).
Thus, each medication \( m_j \) maps to a hierarchical label vector \( y_j = \{y_j^1, y_j^2, \ldots, y_j^K\} \), where \( y_j^i \) is its ATC code at level \( i \).
\subsubsection{Drug–Drug Interaction (DDI)}
\label{subsubsec3}
The DDI knowledge is extracted from the TWOSIDES database\cite{tatonetti2012data}, a comprehensive repository of polypharmacy side effects. The DDI network is represented as a symmetric binary adjacency matrix \(\mathcal{A}_{ddi} \in \{0, 1\}^{|\mathcal{M}|\times |\mathcal{M}|}\), where each entry \((\mathcal{A}_{ddi})_{jk} = 1\) indicates a documented adverse interaction between drugs \(m_j\) and \(m_k\); and \((\mathcal{A}_{ddi})_{jk} = 0\) signifies no known interaction. For medications in the MIMIC-III cohort that lack explicit interaction profiles in TWOSIDES, we set the corresponding entries to 0, indicating no documented significant interaction. This matrix serves as a safety constraint to penalize recommendations with high interaction risks, thereby enhancing the clinical validity of the model.
\subsubsection{Core Task: Hierarchy and Knowledge-Aware Medication Recommendation}
\label{subsubsec4}
Given a patient's complete EHR sequence \(\mathcal{P}^{(i)}\) up to the \(T\)-th visit, together with the medication set \(\mathcal{M}\), the ATC hierarchy, and the DDI matrix \(\mathcal{A}_{ddi}\), the goal is to predict the set of medications \(\mathcal{M}_T \subseteq \mathcal{M}\) to be recommended. The prediction must integrate molecular drug knowledge and strictly adhere to ATC hierarchical logic to ensure clinical coherence.

\begin{figure}[t]
\centering
\includegraphics[width=1\textwidth]{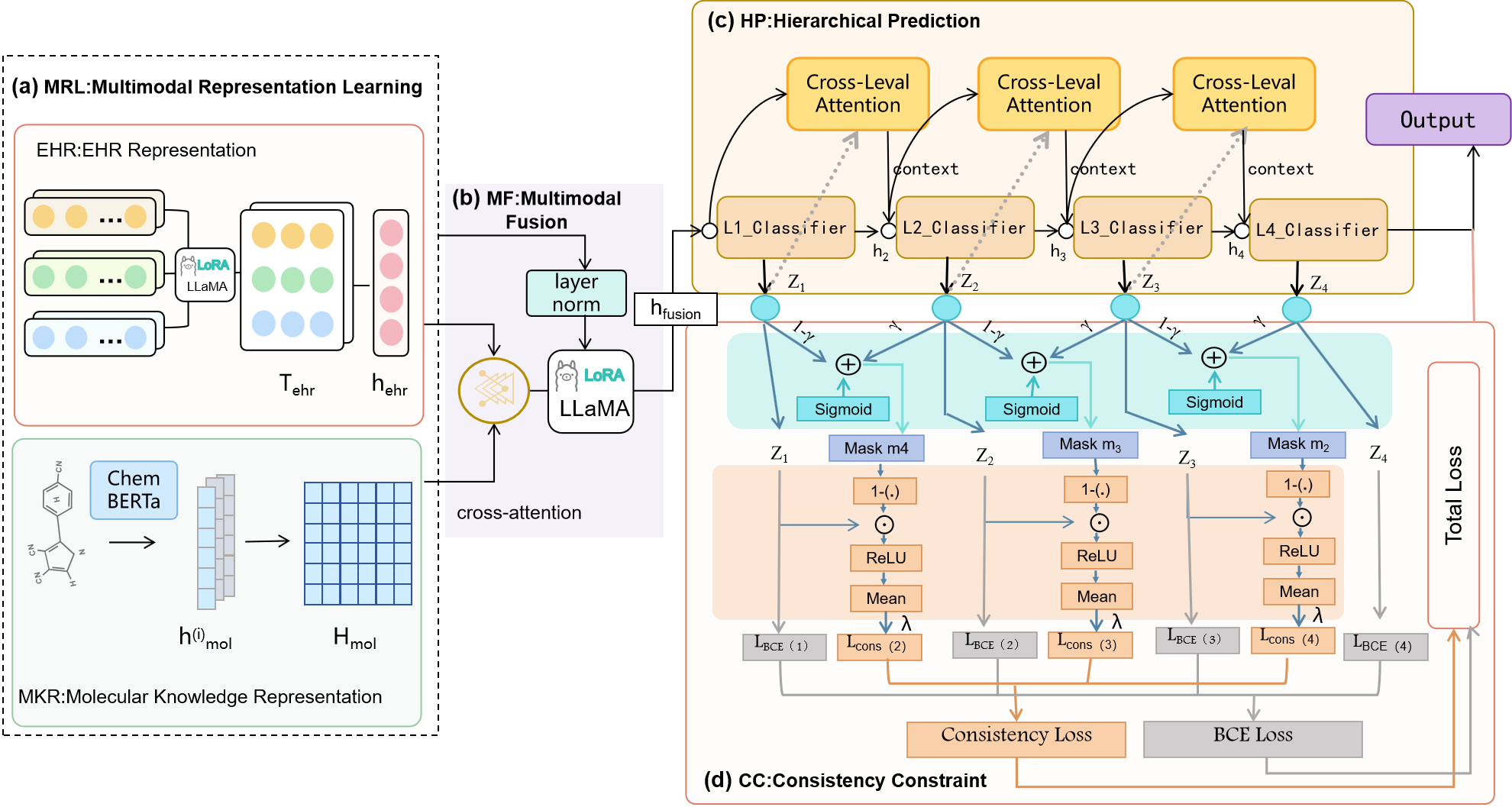}
\caption{The workflow of the HADRec framework: (a) Multimodal representation learning employs LLaMA-7B and ChemBERTa to extract patient clinical profiles and drug molecular features, respectively. (b) Multimodal fusion performs deep interactions between patient states and drug molecular knowledge via a cross-attention mechanism. (c) The hierarchical prediction module constructs a four-level predictor to generate drug recommendations following the top-down ATC taxonomy. (d) The consistency constraint module incorporates a consistency loss function to enforce clinical logicality in the recommendations.}\label{fig1}
\end{figure}

\subsection{Overall Framework}
\label{subsec2}
This section details HADRec, a hierarchy-aware drug recommendation framework which integrates molecular knowledge with EHRs. The core idea of HADRec is to embed fine-grained molecular semantics into patient representations through multimodal fusion, thereby equipping the model with pharmacological context for mechanistic reasoning. Simultaneously, a hierarchical prediction architecture coupled with consistency constraints ensure recommendations are both pharmacologically sound and aligned with the ATC taxonomy’s structural logic.

As illustrated in Figure \ref{fig1}, HADRec consists of four main modules: (a) Multimodal representation learning. A LoRA-adapted LLaMA-7B model encodes patient clinical texts to produce high-quality patient representations. In parallel, ChemBERTa encodes drug SMILES strings to construct a global molecular knowledge base. (b) Multimodal fusion. A cross-attention mechanism enables deep integration between patient clinical states and molecular drug features, outputting a pharmacology-enriched patient representation. (c) Hierarchical prediction. A four-level prediction architecture is constructed to follow the ATC taxonomy in a strict top-down fashion. This design mimics the clinical decision pathway from diagnosis to therapeutic selection, ensuring structural coherence in the recommended drug set. (d) Consistency constraint. Beyond standard classification losses, a novel hierarchy-consistency loss is introduced to enforce strict adherence to ATC-level dependencies. This mechanism guarantees that the final recommendation is clinically credible and logically consistent.

\subsection{Multimodal Representation Learning Module}
\label{subsec3}
This module provides high-quality inputs representations for downstream prediction by capturing the complex relationship between patient clinical states and the intrinsic chemical properties of medications, thereby laying the foundation for accurate and reliable recommendations.
\subsubsection{Patient Clinical Representation}
\label{subsubsec5}
\begin{figure}[H]
\centering
\includegraphics[width=1\textwidth]{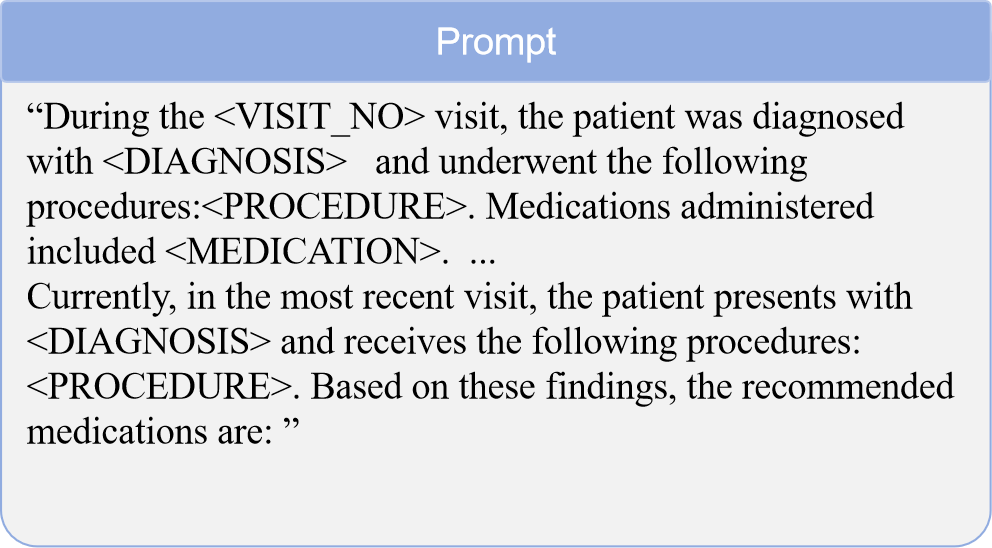}
\caption{Illustration of the prompting template used for model input. The placeholders <...> are populated with patient EHR data.}\label{fig2}
\end{figure}

To precisely characterize a patient’s health states and treatment context, we construct a comprehensive clinical text \( T_{ehr} \). For the \( T \)-th visit, we integrate historical visits $\mathcal{H}_T^{(i)}$ (all preceding visits) along with the current visit $\tilde{v}_T^{(i)}$  into a coherent natural-language description using a structured template (see Figure \ref{fig2}). This templating strategy ensures that only chronologically prior information and current diagnostic evidence are accessible to the model, strictly preventing any temporal data leakage from the target medication set \(\mathcal{M}_{t}^{(i)}\).

We then employ a pretrained large language model as the EHR encoder. Its deep Transformer architecture captures complex medical terminology and longitudinal dependencies in  \( T_{ehr} \), outputting a fixed-dimensional vector \( h_{ehr} \in \mathbb{R}^{dim} \) that reflects high-level patient semantics. This representation serves as an information-rich foundation for subsequent medication prediction.
\begin{equation}
h_{ehr} = LLM(Tokenizer(T_{ehr}))
\end{equation}

\subsubsection{Molecular Knowledge Base Construction}
\label{subsubsec6}
To overcome the limitations of treating medications as abstract identifiers, we incorporate molecular structure information to establish a pharmacologically grounded foundation for model reasoning. We utilize the SMILES string as the standard representation for each drug. A SMILES string employs a linear symbolic notation to describe atom types, chemical bonds, and functional group connectivity, thereby translating complex three-dimensional molecular structures into machine-readable sequences. To extract rich chemical semantics encoded with these SMILES strings, we employ the pretrained ChemBERTa model as the molecular encoder. For every medication \( m_j \) in the drug set, its SMILES strings \( s_j \) is encoded into a high-dimensional molecular embedding \( h_{mol}^{(j)} \). The embeddings of all drugs collectively form a global molecular knowledge matrix \( \mathcal{H}_{mol} \in \mathbb{R}^{N_{drug} \times d_{dim}} \), where \( N_{drug} \) denotes the total number of candidate drugs.
\begin{equation}
h_{mol}^{(j)} = proj \left( ChemBERTa \left( Tokenizer(s_j) \right) \right)
\end{equation}

This encoding step is precomputed. The resulting knowledge matrix \( \mathcal{H}_{mol} \) serves as a static, queryable global resource during both training and inference, supplying immediate molecular features that support the downstream cross-modal fusion process.
\subsection{Multimodal Fusion Module}
\label{subsec4}
This module bridges the persistent gap between clinical and molecular semantics by establishing deep pharmacological grounding within patient representations. As its core, a cross-attention mechanism selectively injects knowledge form the global drug repository into the patient’s clinical profile, yielding a unified representation enriched with mechanistic drug information. Crucially, this design enables adaptive, data-dependent weighting over the medication space, mitigating the risk of the model defaulting to frequently prescribed drug categories by dynamically prioritizing clinically relevant but statistically underrepresented molecular patterns.Specifically, after obtaining the unimodal representations, the patient embedding \( h_{ehr} \) serves as the Query. The global molecular knowledge matrix \( \mathcal{H}_{mol} \), which contains the embeddings of all drugs, serves as both the Key and the Value. The model computes the similarity between the patient embedding \( h_{ehr} \) and each drug vector \( h^{(j)}_{mol} \) to derive an attention distribution \( a \). This distribution quantifies the alignment strength between the patient's current clinical state and the underlying pharmacological mechanisms of each drug, effectively acting as a soft selection over the drug space:
\begin{equation}
\alpha = Softmax \left( \frac{(h_{ehr} W_Q)(\mathcal{H}_{mol} W_K)^T}{\sqrt{d_k}} \right) 
\end{equation}
where \( W_Q \) and \( W_K \) are trainable projection matrices, and \( d_k \) is the dimensionality of the projected Key vectors. The attention weights  are then used to aggregate the molecular knowledge \( \mathcal{H}_{mol} \) through a weighted sum, producing a patient-specific pharmacological context vector \( C_{drug} \):
\begin{equation}
C_{drug} = \alpha (\mathcal{H}_{mol} W_v) 
\end{equation}

Finally, this context is integrated with the original patient embedding through a residual connection, accompanied by a learnable scaling parameter \(\lambda\) and layer normalization, yielding the final molecule-aware patient representation \(h_{fusion}\).
\begin{equation}
h_{fusion} = LayerNorm(h_{ehr} + \lambda C_{drug}) 
\end{equation}

The resulting ``molecule-aware'' representation \(h_{fusion}\) serves as the unified input to the subsequent hierarchical prediction module, jointly encoding the patient’s individualized clinical profile and the most pharmacologically relevant molecular knowledge.
\subsection{Hierarchical Prediction Module}
\label{subsec5}
To produce recommendations that are both accurate and clinically coherent, this module moves beyond conventional flat prediction schemes by adopting a top-down hierarchical architecture. The design mirrors the tree structure of the ATC coding system, thereby emulating the clinical reasoning process that progresses from broad therapeutic categories toward specific treatment decisions.

We construct a four-stage hierarchical framework, with each stage corresponding to one level of the ATC taxonomy (L1–L4, e.g., A → A0 → A01 → A01A). A chain-structured cross-level attention mechanism enables iterative refinement across layers, allowing the model to transition smoothly from high-level therapeutic intent to precise drug selection. The prediction begins at the top level (L1), where the fused patient representation \( h_{fusion} \) is passed into the L1 classifier to generate multi-label logits \( Z_1 \) over all L1 categories:
\begin{equation}
Z_1 = W_1 h_{fusion} + b_1
\end{equation}

For the subsequent level (L2), we introduce a cross-level information-transfer mechanism to ensure predictions are contextually informed rather than made in isolation. Prior to computing the L2 outputs, a cross-layer attention module uses the L1 logits \( Z_1 \) to refine the input representation for the current layer. A notable feature of this design is that the attention mechanism operates directly on the weight matrix \( W_1 \) of the L1 classifier, which uses the L1 logits \( Z_1 \) as attention weights, instead of manipulating hidden states. 
\begin{equation}
ctx_2 = CrossAttn(h_{fusion} Z_1, W_1)
\end{equation}
\begin{equation}
h_2 = LayerNorm(h_{fusion} + ctx_2)
\end{equation}
\begin{equation}
Z_2 = W_2 h_2 + b_2
\end{equation}
Here, \( Z_1 \) acts as a structured prior, guiding the lower-level predictor toward clinically admissible subcategories. The same refinement strategy is applied recursively to deeper layers: L3 predictions are conditioned on L2 outputs, and L4 predictions on L3 outputs, forming a coherent chain of progressively focused inferences:
\begin{equation}
h_{i+1} = LayerNorm(h_i + CrossAttn(h_i, Z_i, W_i))
\end{equation}
\begin{equation}
Z_{i+1} = W_{i+1} h_{i+1} + b_{i+1}
\end{equation}

This chained hierarchical structure transforms independent classification tasks into a unified, stepwise reasoning process. By strictly constraining the search space within activated therapeutic paths, the model naturally balances the representation of rare drug categories and prevents defaulting to globally high-frequency labels. It not only improves predictive accuracy but also substantially enhances interpretability by aligning the decisions with clinically recognized therapeutic hierarchies.
\subsection{Consistency Constraint Module}
\label{subsec6}
Although the hierarchical prediction module enforces a top-down information flow, the probability outputs across different levels may still be logical inconsistencies during training. To ensure strict adherence to the ATC dependency structure, we introduce a novel hierarchical consistency loss. This constraint prevents, for example, the prediction of an L3 class like “C09 – Agents acting on the renin–angiotensin system” unless its parent L1 category “C – Cardiovascular system” is activated.

This loss is implemented through a soft-masking mechanism. For each of the L2, L3, and L4 levels, we construct a soft mask that integrates guidance from upper level with the current layer’s own prediction. Taking L2 as an example, its mask \( m_2 \in [0,1]^{N_2} \)  is computed by blending the mean predicted probability of L1 with the predicted probability of L2:
\begin{equation}
m_2 = \gamma \cdot mean(\sigma (Z_1)) + (1 - \gamma) \sigma (Z_2)
\end{equation}
where \(\sigma\) is the sigmoid function and \(\gamma\) is a balancing hyperparameter (set to 0.5 default). This masking strategy acts as a structural filter: it ensures that the loss landscape favors predictions aligned with the taxonomic hierarchy, thereby preventing the model from erroneously defaulting to globally frequent drugs that do not belong to the currently activated therapeutic path. The hierarchical consistency loss \(\mathcal{L}_{consistency}\) penalizes cases in which high activations occur in regions assigned low plausibility by the mask. Formally, it is computed by applying the ReLU activation function to the element-wise product of the mask complement \((1-m_{i})\) and the current-layer logits \(Z_{i}\), then averaging the result:
\begin{equation}
\mathcal{L}_{consistency}=\sum_{i=2}^{4}mean\left(ReLU\big{(}Z_{i}\odot(1-m_{i}) \big{)}\right)
\end{equation}
Where, \(\odot\) denotes element-wise multiplication. This soft constraint avoids the gradient vanishing and instability issues often associated with hard constraints, while acting as an effective hierarchical regularizer. It guides the model to internalize tmitigating class imbalance by penalizing "lazy" high-frequency guesses that violate these dependencies.

The final training objective combines a multi-task classification loss with the consistency loss. For each layer \(i\), a binary cross-entropy loss is applied:
\begin{equation}
\mathcal{L}_{total}=\sum_{i=1}^{4}\mathcal{\lambda}_{i}\cdot\mathcal{L}^{(i)}_{BCE}+ \lambda_{cons}\cdot\mathcal{L}_{consistency}
\end{equation}
where, \(\mathcal{L}^{(i)}_{BCE}\) is the binary cross-entropy loss for level \(i\), \(\mathcal{\lambda}_{i}\) is the corresponding weight, and \(\lambda_{cons}\) controls the contribution of the consistency term. By optimizing this joint objective end-to-end, the framework learns a recommendation strategy that is both accurate and rigorously aligned with pharmacological reasoning and the hierarchical logic of the ATC taxonomy.
\section{Experimental Setup}
\label{sec4}
\subsection{Dataset}
\label{subsec7}
We evaluate the proposed framework using the MIMIC-III (Medical Information Mart for Intensive Care III) database\cite{johnson2016mimic}, a large-scale, public available dataset of de-identified health records. Maintained by the MIT Laboratory for Computational Physiology, MIMIC-III contains comprehensive clinical data for over 40,000 patients admitted to intensive care units at the Beth Israel Deaconess Medical Center. It includes a wide range of multimodal clinical information, such as demographics, vital signs, laboratory measurements, imaging reports, diagnostic codes, procedure codes, and medication administration records. To adapt the data for our hierarchical medication recommendation task, we constructed an analysis cohort by extracting all patient visit with recorded medication orders. Clinical histories were built from sequences of diagnosis and procedure codes, while all administered medications were mapped to their corresponding ATC codes. This processing yielded a structured dataset where each visit is associated with a set of clinical codes and a set of ATC-coded medications, suitable for evaluating knowledge-aware and hierarchy-consistent recommendation models. The resulting dataset was randomly partitioned at the patient level into training, validation, and test sets with a ratio of 8:1:1. Table \ref{table1} summarizes the key statistics of the processed dataset.

\begin{table}[h]
\centering
\caption{Summary statistics of the processed dataset derived from MIMIC-III.}
\begin{tabular}{p{3.5cm} p{6cm} r}
\hline
\textbf{Category} & \textbf{Item} & \textbf{Count} \\
\hline
Patients & Total patients & 6350 \\
& Average visits patient & 2.37 \\
\hline
Clinical Events & Unique diagnosis codes & 1958 \\
& Unique procedure codes & 1430 \\
\hline
Medications & Unique medications (ATC-L4) & 110 \\
& Average medications per visit & 19.57 \\
\hline
ATC Hierarchy & Categories at Level 1 & 14 \\
& Categories at Level 2 & 17 \\
& Categories at Level 3 & 59 \\
& Categories at Level 4 & 110 \\
\hline
Data Split &  Training patients &  5080 \\
 &  Validation patients &  635 \\
 &  Test patients &  635 \\
\hline
\end{tabular}
\label{table1}
\end{table}

\subsection{Historical Context Window Selection}
\label{subsec17}
HADRec processes patient EHRs as a temporal sequence of clinical visits. For each target visit, we include up to three most recent prior visits as historical context; patients with fewer than three prior visits contribute all available records. The choice of a three-visit window is empirically justified by the analysis in Figure \ref{fig5} (\ref{app1}), which is based on the filtered MIMIC-III cohort (N=6,350).

As shown in Figure \ref{fig5}, 89.0\% of patients have three or fewer hospitalizations. For these patients, a three-visit window preserves their complete longitudinal history without any truncation. For the remaining 11.0\% of patients with longer histories, the window still retains on average 63.1\% of their total visits. Critically, the discarded early visits are on average 2.4 years older than the target admission, whereas the overall median inter-visit interval is only about four months, indicating that older records have limited prognostic value for predicting current medications. Furthermore, increasing the window size beyond three yields rapidly diminishing marginal gains in patient coverage (e.g., from W=3 to W=4, coverage increases by only +5.9\%), yet substantially increases sequence length and computational cost.

Therefore, selecting W=3 strikes an optimal balance between longitudinal information coverage, clinical relevance, and computational efficiency.

\subsection{Baseline Models}
\label{subsec8}
\begin{table}[htbp]
\centering
\small
\caption{Categorization and key characteristics of the baseline models.}
\label{table2}
\renewcommand{\arraystretch}{1.3}
\begin{tabular}{@{}>{\raggedright}p{2.3cm}>{\raggedright}p{2.2cm}>{\raggedright}p{3.6cm}>{\raggedright}p{3.3cm}>{\centering\arraybackslash}p{1.2cm}@{}}
\toprule
\textbf{Category} & \textbf{Model} & \textbf{Key-Technique} & \textbf{Drug-Knowledge Used} & \textbf{Use ATC?} \\
\midrule

\multirow{4}{*}{\parbox{2.3cm}{\centering Knowledge-Graph \& EHR-based}} 
& GAMENet\cite{shang2019gamenet} & Memory-Network \& GNN & Knowledge-Graph (KG) & $\times$ \\
& SafeDrug\cite{yang2021safedrug} & DDI-controllable-Loss & Molecular-Graph & $\times$ \\
& RASNet\cite{zhu2024rasnet} & Reverse-Time Attention & -- & $\times$ \\
& EGNeT\cite{nguyen2025egnet} & GNN \& Explainable Module & External-Medical-KG & $\times$ \\
\midrule

\multirow{3}{*}{\parbox{2.3cm}{\centering Molecular-Structure-Aware}}
& DEPOT\cite{zhao2024enhancing} & Motif-Tree \& Graph-Transformer & Molecular-Motifs & $\times$ \\
& MoleRec\cite{yang2023molerec} & Substructure Interaction \& Annealed Loss & Molecular Substructures & $\times$ \\
& AMGNet\cite{li2025amgnet} & Multi-Graph Representation Learning & Drug Molecular Structure & $\times$ \\
\midrule

\parbox{2.2cm}{\centering Hierarchy-Aware}
& StarMed\cite{li2024stratmed} & Dual-Attribute-Graph Network & -- & $\checkmark$ \\
\bottomrule
\end{tabular}
\end{table}
To thoroughly evaluate HADRec, we compare it against a comprehensive set of state-of-the-art baselines from key methodological strands in medication recommendation. These models are categorized into three primary groups based on their core technical paradigms: (1) models that integrate external knowledge graphs with EHRs, (2) models that incorporate drug molecular structures, and (3) models that attempt to leverage hierarchical information. A summary is provided in Table \ref{table2}

(1).Knowledge-Graph \& EHR-based Models This category of methods focuses on enriching longitudinal patient representations with structured external knowledge, primarily to address safety and generalization.
 
 \hspace*{2em}GAMENet\cite{shang2019gamenet} is a hybrid framework that integrates drug knowledge graphs with a patient’s medication history, leveraging graph neural network (GNN) and memory networks to generate personalized and safety-aware recommendations.

 \hspace*{2em}SafeDrug\cite{yang2021safedrug} captures drug characteristics at the molecule and substructure levels. It employs a differentiable, DDI-controllable loss function to explicitly penalize and mitigate adverse drug interaction risks during training. 

 \hspace*{2em}RASNet\cite{zhu2024rasnet} uses a recurrent aggregation network to retrieve historical visits most relevant to the current clinical state, and introduces an exponential controller to dynamically modulate DDI-related penalties.

 \hspace*{2em}EGNeT\cite{nguyen2025egnet} enhances both performance and interpretability by integrating external medical knowledge graphs with patient EHRs via a GNN, and it incorporates similarity-based explanation modules. 

(2).Molecular-Structure Aware Models. These models directly incorporate the molecular structures of drugs to ground recommendations in pharmacology, representing a significant advance over abstract drug tokens.

 \hspace*{2em}DEPOT\cite{zhao2024enhancing} decomposes each drug into a semantic motif tree and employs a structure-aware graph Transformer to model cooperative interactions among motifs. It further refines recommendations with adaptive motif-disease relationship modeling.

 \hspace*{2em}MoleRec\cite{yang2023molerec} learns the interactions between molecular substructures and their associations with patient health states. It introduces an annealed weighting strategy to adaptively regulate the DDI loss term for safer combination recommendations.

 \hspace*{2em}AMGNet\cite{li2025amgnet} integrates multi-graph representation learning including drug molecular structures, EHR co-occurrence graphs, and DDI graphs, employing a transformer encoder-decoder for patient longitudinal modeling and contrastive learning between EHR and DDI graphs to balance recommendation accuracy with safety.

(3).Hierarchy-Aware Model. A limited number of works have begun to explore the use of the ATC taxonomy’s structure.

 \hspace*{2em}StartMed\cite{li2024stratmed} proposes a correlation-driven, hierarchical architecture to improve learning under sparse data conditions. It constructs a dual-attribute graph within a single hierarchical level to jointly model medication safety and efficacy. However, it does not enforce a strict, multi-level top-down prediction logic.

This selection of baselines allows us to rigorously evaluate HADRec’s contributions: its deep cross-modal fusion of molecular knowledge against Category 2 models, and its strict hierarchical consistency against the flat or weakly constrained approaches of Categories 1 and 3.

\subsection{Evaluation Metrics}
\label{subsec9}
To comprehensively assess model performance from both predictive accuracy and clinical safety perspectives, we employ the following five established metrics: the Jaccard similarity coefficient, F1-score, Precision-Recall Area Under the Curve (PR-AUC), Drug-Drug Interaction (DDI) Rate, and the Average Number of Medications per patient (Avg\_med).

\textbf{Jaccard}: This metric calculates the size of the intersection divided by the size of the union of the predicted and ground-truth medication sets. It provides a direct measure of the overall set-level similarity.

\textbf{F1 Score}: As the harmonic mean of precision and recall, the F1-score balances the model’s ability to correctly identify relevant medications (precision) against its ability to retrieve all necessary ones (recall).

\textbf{PR-AUC}: The area under the precision–recall curve is particularly informative for our multi-label classification task with inherent class imbalance. It summarizes the trade-off between precision and recall across all probability thresholds.

\textbf{DDI Rate}: Defined as the proportion of predicted drug pairs that have known adverse interactions, this rate is a critical safety indicator. A lower DDI rate signifies a safer recommendation.

\textbf{Avg\_med}: This metric reports the average count of medications predicted per patient visit, offering insight into the model’s prescribing tendency and helping to contextualize other metrics.

\begin{table}[htbp]
\centering
\caption{Hyperparameter settings of the HADRec framework}
\small  
\begin{tabular}{@{}p{4cm}p{2cm}p{5cm}p{3cm}@{}}
\toprule
\textbf{Category} & \textbf{Notation} & \textbf{Description} & \textbf{Value} \\
\midrule

\multirow{2}{*}{Model Architecture} 
& $d_{dim}$ & Hidden layer dimension & 4096 \\
& $d_{mol}$ & Molecular embedding dimension & 768 \\
\addlinespace

\multirow{2}{*}{Attention Mechanism} 
& $\lambda$ & Residual connection scaling factor & 0.7 \\
& $\tau_{cross}$ & Soft masking threshold for cross-attention & 0.1 \\
\addlinespace

\multirow{2}{*}{Loss Weighting} 
& $(\lambda_i)_{i=1}^{4}$ & Weights for hierarchical BCE loss (L1-L4) & 0.4, 0.4, 1.0, 1.5 \\
& $\lambda_{cons}$ & Global weight for consistency constraint loss & 0.01 \\
\addlinespace

\multirow{3}{*}{Consistency Constraints} 
& $w_{cons\_L2}$ & Weight for Level-2 consistency constraint & 1.0 \\
& $w_{cons\_L3}$ & Weight for Level-3 consistency constraint & 1.5 \\
& $w_{cons\_L4}$ & Weight for Level-4 consistency constraint & 2.0 \\
\bottomrule
\end{tabular}
\label{table3}
\end{table}

\subsection{Implementation Details}
\label{subsec10}
This section outlines the key implementation configurations and hyperparameter settings for HADRec. We adopt LLaMA-7B and ChemBERTa as the foundational encoders for clinical text and drug SMILES strings, respectively. The model is trained using the AdamW optimizer with a learning rate of $2 \times 10^{-4}$. The maximum input sequence length is set to 2048 tokens, and training proceeds for up to 5000 steps. To improve computational efficiency, we enable FP16 mixed-precision training. During inference, predictions are binarized with a threshold of 0.35. All experiments are conducted on a single NVIDIA RTX 3090 GPU. The primary hyperparameters are summarized in Table \ref{table3} .

\section{Results and Discussion}
\label{sec5}
\subsection{Overall Performance Comparison}
\label{subsec11}
To comprehensively assess the effectiveness of the proposed HADRec framework, we benchmarked it against a set of state-of-the-art baselines outlined in Section \ref{subsec8}. As summarized in Table \ref{table4}, HADRec achieves the best performance on all three core accuracy metrics, obtaining the highest Jaccard (0.5517), F1-score (0.7010), and PR-AUC (0.7994). This consistent superiority holds even when compared to strong baselines that integrate molecular structures (e.g., EGNet) or external knowledge graphs (e.g., RASNet). In terms of clinical safety and prescription rationality, HADRec also demonstrates compelling advantages. It maintains a competitively low DDI rate (0.0774), Notably, HADRec predicts a relatively small number of medications per patient (Avg\_Med = 20.51), significantly lower than most competitors. This indicates that its accuracy gains stem from recommending more precise and necessary drug combinations, rather than from increasing the prescription size.

\begin{table}[htbp]
\centering
\footnotesize  
\setlength{\tabcolsep}{3.5pt}  
\caption{Performance comparison between HADRec and baseline models. Results are reported as mean ± standard deviation across multiple runs. The best performances are highlighted in bold.}
\label{table4}

\begin{tabular}{@{}>{\raggedright}p{2.3cm} l c c c c c@{}}
\toprule
\textbf{Category} & \textbf{Model} & \textbf{Jaccard $\uparrow$} & \textbf{F1 $\uparrow$} & \textbf{PR-AUC $\uparrow$} & \textbf{DDI $\downarrow$} & \textbf{Avg.Med $\downarrow$} \\
\midrule

\multirow{4}{*}{\parbox{2.5cm}{Knowledge-Graph \& EHR-based}}  
& GAMENet\cite{shang2019gamenet} & 0.5067±0.0025 & 0.6626±0.0025 & 0.7631±0.0030 & 0.0864±0.0006 & 27.21±0.11 \\
& SafeDrug\cite{yang2021safedrug} & 0.5213±0.0030 & 0.6768±0.0027 & 0.7647±0.0025 & \textbf{0.0589±0.0005} & \textbf{19.92±0.16} \\
& RASNet\cite{zhu2024rasnet} & 0.5401±0.0021 & 0.6931±0.0019 & 0.7882±0.0025 & 0.0599±0.0009 & 21.04±0.15 \\
& EGNeT\cite{nguyen2025egnet} & 0.5431±0.001 & 0.6955±0.0007 & 0.7842±0.0023 & 0.0693±0.0012 & 23.33±0.01 \\
\midrule

\multirow{3}{*}{\parbox{2.5cm}{Molecular-Structure-Aware}}
& DEPOT\cite{zhao2024enhancing} & 0.5352±0.0022 & 0.6890±0.003 & 0.7816±0.0035 & 0.0684±0.0007 & 20.62±0.19 \\
& MoleRec\cite{yang2023molerec} & 0.5301±0.0025 & 0.6841±0.0022 & 0.7748±0.0022 & 0.0756±0.0006 & 22.22±0.17 \\
& AMGNet\cite{li2025amgnet} & 0.5416±0.0023 & 0.6887±0.0027 & 0.7824±0.0016 & 0.0590±0.0011 & 20.29±0.12 \\
\midrule

Hierarchy-Aware
& StratMed\cite{li2024stratmed} & 0.5321±0.0035 & 0.6861±0.0034 & 0.7779±0.0043 & 0.0642±0.0005 & 20.53±0.17 \\
\midrule

\multirow{1}{*}{\parbox{2.5cm}{Ours}}
& HADRec & \textbf{0.5517±0.0059} & \textbf{0.7010±0.0054} & \textbf{0.7994±0.0057} & 0.0774±0.011 & 20.51±0.27 \\
\bottomrule
\end{tabular}

\end{table}

Overall, the results demonstrate that HADRec achieves superior accuracy, promotes prescription conciseness, and maintains a high safety standard. This balanced excellence confirms that the framework successfully integrates molecular knowledge and hierarchical constraints to advance the state of the art in medication recommendation.
\subsection{Ablation Study}
\label{subsec12}
To quantitatively evaluate the contribution of each core component in HADRec, we conducted a series of ablation experiments. The results, summarized in Table \ref{table5}, analyze the impact of three key modules: the molecular knowledge representation (MKR), the hierarchical prediction architecture (HP), and the consistency loss (CL). 

Removing the MKR module leads to a marked drop in all accuracy metrics compared to the full model (cf. Row 4 vs. Row 5). This confirms that fine-grained molecular semantics provides essential pharmacological prior and enhance the model’s ability to align patient states with drug mechanisms. Removing the CL module while keeping HP also results in degraded performance (cf. Row 3 vs. Row 5). This demonstrates that the explicit hierarchical constraints are crucial for guiding finer-grained predictions and for mitigating logical inconsistencies across ATC levels. Furthermore, removing both the HP and CL modules—thereby reverting to a flat prediction structure while retaining molecular knowledge—results in the most substantial performance decrease across all metrics (cf. Row 2 vs. Row 5). This underscores that hierarchical reasoning framework (HP + CL) is indispensable for the structured, progressive reasoning that supports more reliable and interpretable clinical decision-making.

These findings collectively demonstrate that each component contributes uniquely to the overall model performance, and that their integrated design is fundamental to HADRec’s effectiveness. The ablation study provides clear mechanistic support for the holistic architecture of our proposed framework.

\begin{table}[h]
\centering
\caption{Ablation study of key components of HADRec}

\begin{tabular}{l l l|l l l}
\hline
MKR & HP & CL & Jaccard  $\uparrow$& F1  $\uparrow$& PR-AUC  $\uparrow$\\
\hline
$\times$ & $\times$  & $\times$  & 0.5397±0.0063 & 0.6904±0.0063 & 0.7946±0.0060 \\
$\checkmark$ &  $\times$ &  $\times$ & 0.5405±0.0090 & 0.6915±0.0087 & 0.7944±0.0066 \\
$\checkmark$ &  $\checkmark$ & $\times$ & 0.5472±0.0051 & 0.6967±0.0050 & 0.7986±0.0052 \\
 $\times$ & $\checkmark$ & $\checkmark$ & 0.5450±0.0054 & 0.6950±0.0053 & 0.7971±0.0049 \\
$\checkmark$ & $\checkmark$ & $\checkmark$  & \textbf{0.5517±0.0059} & \textbf{0.7010±0.0054} & \textbf{0.7994±0.0057} \\
\hline
\end{tabular}

\label{table5}

\raggedright Note: ‘$\times$’ indicates the component is disabled; ‘$\checkmark$’ indicates enabled. The configuration ‘MKR=$\times$, HP=$\times$, and CL=$\times$ ’ represents predictions made by the base LoRA-adapted LLaMA-7B model without any proposed enhancements. Results are reported as mean ± standard deviation across multiple runs. The best result among all configurations is highlighted in bold. $\uparrow$ indicates higher is better.
\end{table}

\subsection{Influence of Hierarchical Labels on Final-Level Prediction}
\label{subsec13}
\begin{figure}[htbp]
\centering
\includegraphics[width=1\textwidth]{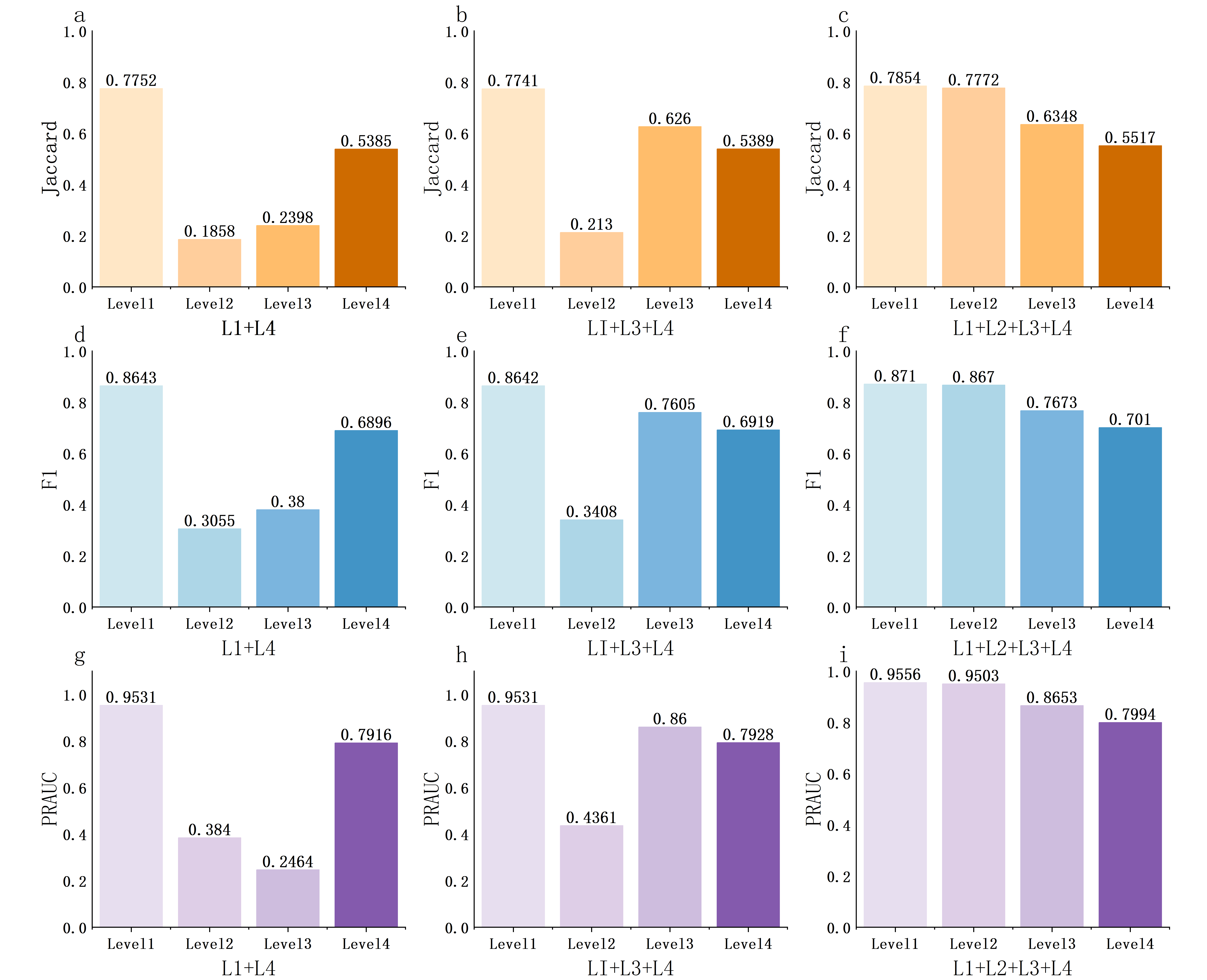}
\caption{Performance comparison demonstrates the superiority of full hierarchical supervision. The figure shows Jaccard (a-c), F1-score (d-f), and PR-AUC (g-i) across four ATC levels (L1-L4) under three supervision settings: L1+L4 (a, d, g), L1+L3+L4 (b, e, h), and full supervision L1+L2+L3+L4 (c, f, i).}\label{fig3}
\end{figure}
To investigate the necessity of supervision at each hierarchical level, we designed a comparative experiment with three loss configurations: L1 + L4 (supervision only at the first and final levels), L1 + L3 + L4 (supervision removed at the second level), and L1 + L2 + L3 + L4 (full hierarchical supervision ). The performance across all four levels (L1–L4) under these settings is comprehensively visualized in Figure 3 for Jaccard, F1, and PR-AUC metrics.

Results consistently demonstrate that full hierarchical supervision yields the strongest overall performance. For instance, at the final level (L4), it achieves a Jaccard of 0.5517 (Figure \ref{fig3}c), an F1-score of 0.7010 (Figure \ref{fig3}f), and a PR-AUC of 0.7994 (Figure \ref{fig3}i). Critically, removing supervision from any intermediate level triggers a cascade degradation effect. This is most severe at the directly unsupervised layer. For example, without L2 supervision, the Jaccard at L2 plummets to 0.2130 (Figure \ref{fig3}b), compared to 0.7772 under full supervision (Figure \ref{fig3}c). When supervision is further reduced (L1+L4), L2 performance deteriorates further to 0.1858 (Figure \ref{fig3}a). This representation collapse propagates downstream, compromising final-layer predictions. Removing L2 supervision alone (L1+L3+L4) causes the final L4 Jaccard to drop from 0.5517 to 0.5389, with corresponding declines in F1 and PR-AUC. Identical performance patterns are observed for F1-score and PR-AUC (Figures \ref{fig3}d-i), confirming the robustness of this cascade effect.

The performance decline under partial supervision can be attributed to a disrupted learning process in which the model fails to develop coherent intermediate representations, the top-down flow of prior guidance is weakened, and gradient optimization becomes disproportionately focused on the final layer. Consequently, the regularization benefit inherent in joint multi-level training is diminished, impairing generalization. In summary, these findings provide conclusive evidence that intermediate ATC-level labels are not redundant. They constitute an essential scaffold for hierarchical reasoning. Complete supervisory guidance ensures robust representation learning, effective prior propagation, and ultimately, superior and more reliable final-layer prediction accuracy.

\subsection{Visualization of Hierarchical Attention}
\label{subsec14}
\begin{figure}[!htbp]
\centering
\includegraphics[width=0.85\textwidth]{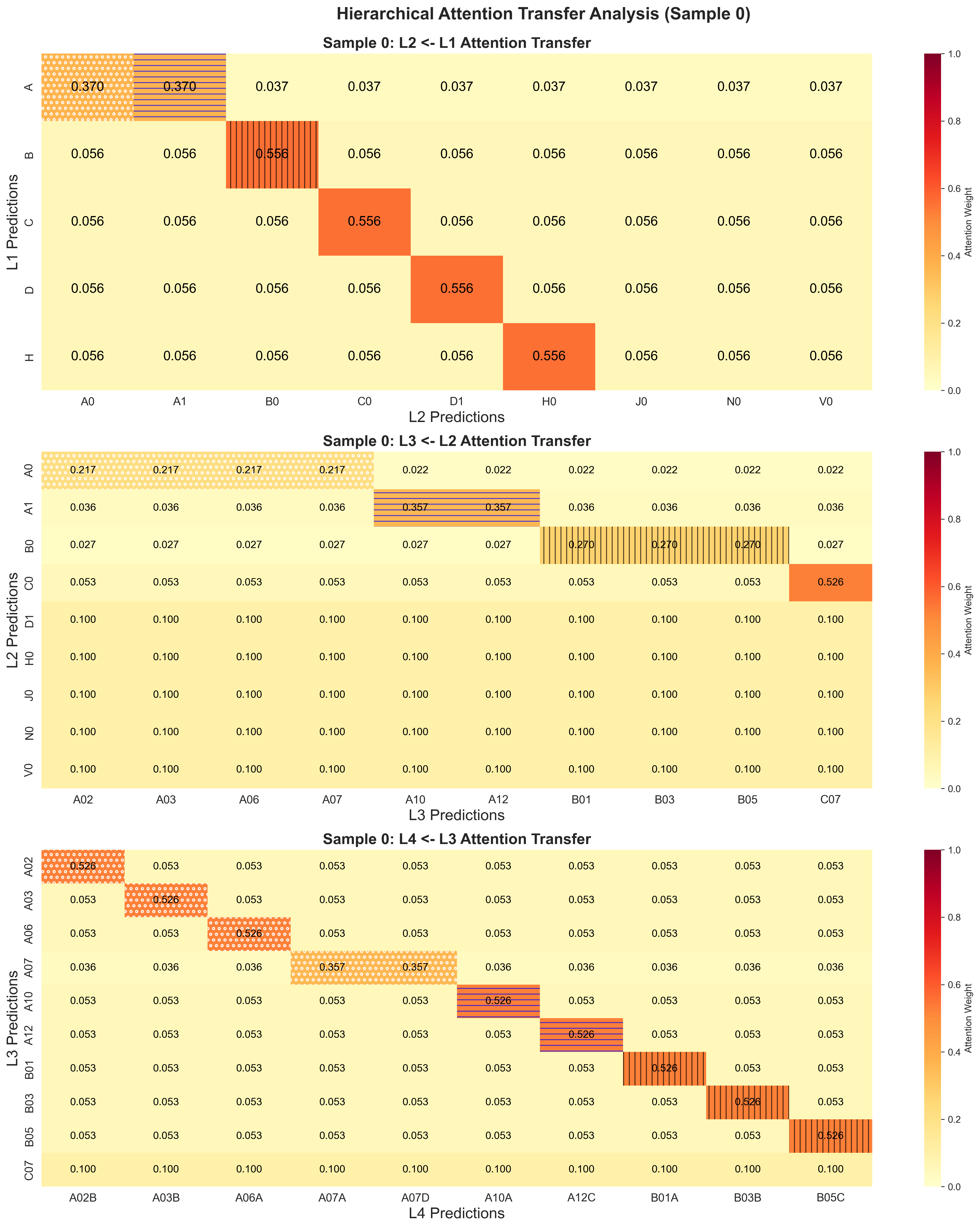}
\caption{Hierarchical attention distribution for a representative patient case. The heatmap visualizes the attention weights across Levels 1 to 4 for the ten drugs with the highest predicted probabilities at Level 4. Darker shading indicates stronger attention, clearly revealing the structured “parent-to-child” information flow constrained by the ATC hierarchy. Cells sharing the same background pattern correspond to drugs belonging to the same therapeutic class, highlighting consistent information transfer within hierarchical groups. Specifically, dotted (·), horizontal (—), and vertical (|) patterns denote A0-, A1-, and B0-level classes, respectively.}\label{fig4}
\end{figure}

To gain deeper insight into the model’s internal reasoning process, we visualize the cross-level attention weights for representative patient cases. The resulting heatmaps (Figure \ref{fig4}) reveal how attention is dynamically routed across the ATC hierarchy during prediction.

Across all layers, high-weight regions consistently cluster within the valid child groups of the corresponding parent categories. For instance, an L1 category routes most of its attention to its legitimate L2 children (e.g., A0/A1). These, in turn, concentrate on their associated L3 subgroups, with L3 further propagating attention to specific L4 categories. This reveals a clear “parent-to-child” aggregation pattern, indicating that the model has internally formed stable hierarchical semantic alignment.

Moreover, strong activation at higher levels leads to more focused attention at lower levels—showing that reliable high-level predictions reinforce fine-grained reasoning. This structured propagation demonstrates that the consistency loss not only constrains the output space but also reshapes information flow, steering the model away from flat multi-label prediction toward coherent, path-dependent hierarchical inference.

Importantly, attention concentration aligns well with the medical semantics of the ATC taxonomy. The model prioritizes subcategories belonging to the same therapeutic domain or chemical family, further validating that consistency constraints enhance medical logic and interpretability.

\subsection{Threshold Sensitivity Analysis and Optimization}
\label{subsec15}
To evaluate the robustness of HADRec and determine the optimal decision boundary for multi-label medication recommendation, we conduct a comprehensive threshold sensitivity analysis by varying the global threshold from 0.1 to 0.9. These results are compared against an optimized threshold calibrated on the validation set and a class-wise dynamic thresholding strategy.

As shown in Table \ref{table6}, the model exhibits a classic precision-recall trade-off as the threshold increases. Specifically, lower thresholds yield high recall (up to 0.9184) but suffer from limited precision, leading to suboptimal F1 scores due to potential over-prescription. Conversely, stringent thresholds significantly enhance precision (up to 0.8765) at the expense of a sharp recall decline, compromising the model's ability to capture necessary therapeutic agents. The optimal balance is achieved at a threshold of 0.35, which yields the highest F1 score of 0.7010 and Jaccard of 0.5517.

\begin{table}[htbp]
\centering
\footnotesize  
\setlength{\tabcolsep}{3.5pt}  
\caption{ Performance comparison of global thresholding (0.1–0.9) and class-wise dynamic thresholding on HADRec. Results are reported as mean ± standard deviation across multiple runs. The best performances are highlighted in bold.}
\label{table6}

\begin{tabular}{@{}>{\raggedright}p{2.3cm} c c c c c@{}}
\toprule
\textbf{method} & \textbf{Threshold} & \textbf{Jaccard $\uparrow$} & \textbf{F1 $\uparrow$} & \textbf{Precision $\uparrow$} & \textbf{Recall $\uparrow$} \\
\midrule

\multirow{10}{*}{\parbox{2.5cm}{Global Threshold}}  
& 0.1 & 0.4578±0.0032 & 0.6186±0.0027 & 0.4789±0.0031 & 0.9184±0.0039 \\
& 0.2 & 0.5253±0.0057 & 0.6791±0.0050 & 0.5848±0.0052 & 0.8443±0.0065 \\
& 0.3 & 0.5491±0.0065 & 0.6992±0.0060 & 0.6589±0.0061 & 0.7741±0.0083 \\
&\textbf{0.35} & \textbf{0.5517±0.0059} & \textbf{0.7010±0.0054} & 0.6891±0.0064 & 0.7403±0.0072 \\
& 0.4 & 0.5470±0.0054 & 0.6969±0.0050 & 0.7126±0.0063 & 0.7069±0.0073 \\
& 0.5 & 0.5275±0.0051 & 0.6789±0.0050 & 0.7595±0.0075 & 0.6349±0.0062 \\
& 0.6 & 0.4950±0.0060 & 0.6487±0.0061 & 0.8027±0.0080 & 0.5636±0.0073 \\
& 0.7 & 0.4445±0.0064 & 0.6002±0.0065 & 0.8437±0.0088 & 0.4841±0.0056 \\
& 0.8 & 0.3641±0.0073 & 0.5161±0.0075 & 0.8765±0.0110 & 0.3834±0.0077 \\
& 0.9 & 0.2421±0.0087 & 0.3654±0.0111 & 0.8669±0.0101 & 0.2467±0.0090 \\
\midrule

Class-Wise Dynamic Thresholding
& \textbf{---} & 0.5219±0.0066 & 0.6760±0.0064 & 0.6112±0.0065 & 0.7878±0.0090\\

\bottomrule
\end{tabular}
\raggedright

\end{table}

Furthermore, we explore a class-wise dynamic thresholding strategy, where thresholds for L4 categories are individually optimized on the validation set by maximizing the F1 score, while the thresholds for the higher levels (L1–L3) remain fixed. This approach yielded comparable performance (F1 = 0.6760, Jaccard = 0.5219), suggesting that while fine-grained thresholding offers additional flexibility, its marginal utility is limited. This indicates that the proposed hierarchy-aware framework already effectively mitigates class imbalance through its intrinsic architectural constraints.
Overall, these results demonstrate that while a uniform threshold can be sensitive to data imbalance, selecting an optimal threshold based on validation performance (0.35 in our case) provides a simple yet effective solution. Moreover, the marginal gains from class-wise thresholding further confirm the robustness of the proposed model in handling imbalanced multi-label prediction.

\subsection{Clinical Case Study and Counterfactual Analysis}
\label{subsec16}
To further investigate the model’s reasoning behavior, we extract a real patient record from the clinical dataset as Case 3, and construct counterfactual variants based on this case by removing key comorbidities. Specifically, Case 1 excludes diabetes mellitus and Case 2 excludes hypertension, while all other conditions remain unchanged, in order to evaluate the model’s response and adjustment of medication recommendations under different clinical condition variations.

We first analyze the impact of diabetes mellitus on antidiabetic drug recommendations (ATC code A10). As expected, when diabetes is absent (Case 1), no A10 medications are recommended. In contrast, when diabetes is present (Case 2 and Case 3), the model consistently recommends A10A (insulins and analogues). This clear and consistent pattern demonstrates that the model effectively captures the associative relationship between diabetes and the need for glucose-lowering therapy. 

\begin{table}[htbp]
\centering
\caption{Counterfactual medication recommendations under selective disease removal.}
\label{table7}

\begin{tabular}{p{3cm} p{6cm} p{6cm}}
\toprule
\textbf{Setting} & \textbf{Disease Profile} & \textbf{Recommended Medication Combination (Top Categories)} \\
\midrule
\textbf{Excluding diabetes} (case 1) & Hypertension, Hyperlipidemia, Chronic kidney disease, Coronary artery disease, Obesity & B01A, C10A, C07A, A01A, A06A, N02B, C09A, B05C, A02B, A12B, A12C \\
\midrule
\textbf{Excluding hypertension} (case 2) & Diabetes mellitus type II, Hyperlipidemia, Chronic kidney disease, Coronary artery disease, Obesity & B01A, C10A, A06A, B05C, A01A, C07A, A12C, A10A, H04A, N02B, A02B, A12B, A04A \\
\midrule
\textbf{Full condition set} (case 3) & Diabetes mellitus type II, Hypertension, Hyperlipidemia, Chronic kidney disease, Coronary artery disease, Obesity & B01A, C10A, A06A, C07A, A01A, B05C, H04A, A10A, A12B, A02B, N02B, C09A \\
\bottomrule
\end{tabular}

\end{table}

Next, we examine antihypertension drugs, including beta-blockers (C07) and ACE inhibitors (C09). When hypertension is present (Case 1 and Case 3), these drug classes are appropriately recommended. When hypertension is removed (Case 2), some antihypertensive drugs (e.g., C07A) are still predicted. However, this behavior is clinically plausible, as beta-blockers are also commonly prescribed for coronary artery disease, which is present in all cases. This indicates that the model does not rely on a single disease label, but instead captures shared therapeutic indications across comorbid conditions in a clinically consistent manner.

Beyond disease-specific therapeutic agents, we observe that certain medications (e.g., A01A, A12B and A06A) are consistently recommended across all scenarios with relatively high confidence. These drugs are not primarily used to treat specific comorbidities, but rather serve as general supportive care commonly administered to critically ill patients (e.g., oral hygiene, gastrointestinal management, routine supplementation). Their stable presence across counterfactual settings suggests that the model can successfully distinguish between disease-specific treatments and routine care.

Overall, these results demonstrate clinically consistent and interpretable medication recommendation behavior in both real-world and counterfactual clinical scenarios under controlled comorbidity perturbations.

\subsection{Multi-center Validation and Transferability Analysis}
\label{subsec18}
To evaluate the generalization capability of the HADRec across heterogeneous clinical environments, we conducted external validation on the MIMIC-IV dataset, a more recent and larger cohort of ICU patients. All data were processed under the same data cleaning, ATC mapping, and evaluation protocol as MIMIC-III. For fair comparison, all baseline models were retrained and evaluated on MIMIC-IV following identical settings. We further performed cross-dataset transfer learning by pre-training on MIMIC-III and fine-turning on MIMIC-IV. The results are summarized in Table \ref{table8}.
\begin{table}[htbp]
\centering
\footnotesize  
\setlength{\tabcolsep}{3.5pt}  

\caption{External validation and cross-dataset transferability evaluation of HADRec on MIMIC-IV.}
\label{table8}
\begin{tabular}{@{}>{\raggedright}p{2.3cm} c c c c c@{}}
\toprule
\textbf{Model} & \textbf{Jaccard$\uparrow$} & \textbf{F1$\uparrow$} & \textbf{PR-AUC$\uparrow$} & \textbf{DDI$\downarrow$} & \textbf{Avg.Med$\downarrow$} \\
\midrule
KNN\cite{shang2019gamenet}       & $0.3978\pm0.0023$ & $0.5473\pm0.0022$ & $0.3917\pm0.0023$ & $0.0815\pm0.0013$ & $12.78\pm0.2203$ \\
LR\cite{luaces2012binary}       & $0.4499\pm0.0033$ & $0.5985\pm0.0027$ & $0.7259\pm0.0024$ & $0.0769\pm0.0008$ & $10.36\pm0.1111$ \\
ECC\cite{read2011classifier}      & $0.4350\pm0.0017$ & $0.5809\pm0.0020$ & $0.7193\pm0.0021$ & $0.0681\pm0.0003$ & $8.76\pm0.2115$ \\
RETAIN\cite{choi2016retain}   & $0.4182\pm0.0020$ & $0.5729\pm0.0016$ & $0.6775\pm0.0012$ & $0.0835\pm0.0004$ & $11.29\pm0.2007$ \\
LEAP\cite{zhang2017leap}     & $0.4123\pm0.0022$ & $0.5661\pm0.0019$ & $0.6026\pm0.0025$ & $0.0667\pm0.0005$ & $12.13\pm0.1530$ \\
GAMENet\cite{shang2019gamenet}   & $0.4594\pm0.0015$ & $0.6126\pm0.0011$ & $0.7229\pm0.0020$ & $0.0803\pm0.0006$ & $18.86\pm0.0002$ \\
MICRON\cite{yang2021change}   & $0.4605\pm0.0023$ & $0.6132\pm0.0021$ & $0.6893\pm0.0022$ & $0.0671\pm0.0003$ & $13.75\pm0.2172$ \\
SafeDrug\cite{yang2021safedrug}  & $0.4643\pm0.0035$ & $0.6165\pm0.0020$ & $0.7051\pm0.0011$ & $0.0672\pm0.0004$ & $14.06\pm0.2615$ \\
COGNet\cite{wu2022conditional}     & $0.4786\pm0.0010$ & $0.6264\pm0.0011$ & $0.6963\pm0.0018$ & $0.0825\pm0.0002$ & $14.71\pm0.1105$ \\
MoleRec\cite{yang2023molerec}  & $0.4721\pm0.0021$ & $0.6241\pm0.0017$ & $0.7145\pm0.0015$ & $0.0724\pm0.0007$ & $14.04\pm0.2203$ \\
DEPOT\cite{zhao2024enhancing}    & $0.4861\pm0.0018$ & $0.6336\pm0.0016$ & $0.7274\pm0.0014$ & $0.0637\pm0.0003$ & $12.77\pm0.1149$ \\
HADRec (MIMIC-IV) & $0.4857\pm0.0055$ & $0.6408\pm0.0055$ & $0.7193\pm0.0136$ & $0.0873\pm0.0214$ & $13.65\pm0.2613$ \\
HADRec (Transfer)  & $0.4307\pm0.0099$ & $0.5901\pm0.008$ & $0.6582\pm0.0115$ & $0.0493\pm0.0178$ & $12.53\pm0.5646$ \\

\bottomrule

\end{tabular}

\end{table}

\subsubsection{External Validation on MIMIC-IV}
\label{subsubsec7}
Despite the distributional differences between MIMIC-III and MIMIC-IV, HADRec maintains competitive performance. Specifically, HADRec trained directly on MIMIC-IV achieves the highest F1-score (0.6408) among all compared methods, while also obtaining competitive Jaccard (0.4857) and PR-AUC (0.7193). This indicates that the ATC-guided hierarchical representations and ChemBERTa-based molecular features generalize effectively across heterogeneous clinical environments.

\subsubsection{Cross-Dataset Transfer Learning}
\label{subsubsec8}
To further evaluate transferability, we conducted a cross-dataset experiment where HADRec was pre-trained on MIMIC-III and fine-tuned on MIMIC-IV. As shown in Table \ref{table8}, the transfer setting leads to a moderate decrease in recommendation accuracy compared with direct MIMIC-IV training, reflecting the distribution shift between datasets. Nevertheless, the transferred model still outperforms several representative baselines while achieving the lowest DDI rate (0.0493) among all methods, suggesting that the learned ATC hierarchy and molecular representations provide stable drug-safety priors across datasets. Overall, these results demonstrate promising generalization and transferability of HADRec.

\subsection{Calibration and Uncertainty Analysis}
\label{subsec19}
To evaluate the reliability of HADRec’s predictive confidence, we conducted model calibration and uncertainty analysis. Given that medical recommendation tasks strictly demand not only high predictive accuracy but also trustworthy probabilistic outputs, we employed Expected Calibration Error (ECE) and Brier Score to assess the probabilistic outputs of the model. Lower values indicate better alignment between predicted probabilities and actual accuracy.

Table \ref{table9} presents the calibration results on MIMIC-III and MIMIC-IV. On MIMIC-III, HADRec achieves an ECE of 0.09 and a Brier Score of 0.12. On MIMIC-IV, the calibration improves further, ECE = 0.04, Brier = 0.06. These values are within the range considered well-calibrated in medical AI literature (ECE < 0.10 is often deemed acceptable). In the cross-dataset transfer setting (MIMIC-III → MIMIC-IV), calibration remains stable, ECE = 0.04, Brier = 0.07. 

These findings suggest that the hierarchical constraint mechanism not only improves predictive accuracy but also mitigates overconfidence, producing trustworthy probability estimates. This is particularly valuable for clinical applications where understanding the model’s confidence is as important as the recommendation itself.
\begin{table}[htbp]
\centering

\caption{Calibration and uncertainty evaluation results.}
\label{table9}
\begin{tabular}{lcc}
\toprule
\textbf{Dataset} & \textbf{ECE} & \textbf{Brier Score} \\
\midrule
MIMIC-III       & 0.09 & 0.12 \\
MIMIC-IV        & 0.04 & 0.06 \\
Transfer (III→IV) & 0.04 & 0.07 \\
\bottomrule
\end{tabular}

\end{table}

\subsection{Efficiency Analysis}
\label{subsec20}
To evaluate the deployment feasibility of the proposed framework, we analyze its computational efficiency on a single NVIDIA RTX3090 GPU (24GB memory). Table \ref{table10} summarizes the inference efficiency. The model achieves an average inference time of 3.14 seconds per sample with a GPU memory footprint of 12.75 GB during inference. While this latency is higher than lightweight models, it is acceptable for offline or semi-real-time clinical decision support (e.g. discharge medication planning or daily round recommendations). Several optimizations could further reduce latency for real-time scenarios, for example, quantizing the LLaMA-7B backbone to INT8 precision, distilling it into a smaller student network, or simply batching multiple patients together. We’ve outlined these possibilities in Section \ref{subsec21} as part of our future work.

\begin{table}[htbp]
\centering

\caption{Inference efficiency and deployment cost analysis.}
\label{table10}
\begin{tabular}{lcccc}
\toprule
\textbf{Model} & \textbf{Backbone} & \textbf{GPU Device} & \textbf{Memory Usage} & \textbf{Avg. Inference Time} \\
\midrule
HADRec & LLaMA-7B & NVIDIA RTX3090 & 12.75GB & 3.14 s/sample \\
\bottomrule
\end{tabular}

\end{table}

\subsection{Limitations and Future Work}
\label{subsec21}
While HADRec demonstrates strong performance in recommendation accuracy, interpretability, and clinical coherence, we acknowledge several limitations that point to future research directions. First, although molecular encodings implicitly capture some pharmacological characteristics, they may not be sufficient for reasoning about complex drug-drug interaction patterns. Future work could incorporate external biomedical knowledge graphs or employ graph neural networks to achieve more systematic DDI modeling. Second, the integration of multimodal encoders (LLaMA-7B, ChemBERTa) with a hierarchical reasoning module introduces nontrivial computational overhead. To support real-time deployment in clinical environments, future studies should explore lighter architectures, model compression, and faster inference strategies.

Beyond these limitations, we plan to extend HADRec in several clinical and technical dimensions. Specifically, we will conduct retrospective and prospective studies in general wards or outpatient settings to assess its impact on patient outcomes, thereby enhancing robustness and generalizability. We also aim to model longitudinal treatment trajectories using time-stamped medication and laboratory data to enable dynamic, adaptive recommendations. To achieve truly personalized medicine, HADRec will incorporate patient-specific factors such as pharmacogenomic profiles and prior treatment responses, with small cohort studies evaluating improvements in drug safety and efficacy. Additionally, we will integrate molecular features with structured medical knowledge graphs using graph neural networks to capture complex drug interactions, assessing the benefits through comparative experiments on interaction prediction and clinical coherence. We will also explore model compression and efficient inference techniques, and conduct deployment trials on ward and outpatient datasets to ensure real-time responsiveness. Overall, HADRec’s mature framework support local retraining and iterative refinement using clinical feedback, offering a feasible and scalable path for deployment across diverse hospital environments.

\section{Conclusion}
\label{sec6}
This study addresses several longstanding challenges in medication recommendation, including the limited expressiveness of symbolic drug representations, the semantic gap between clinical notes and molecular pharmacology, and the absence of ATC-consistent hierarchical logic in predicted prescriptions. To mitigate these issues, we propose HADRec, a hierarchical-aware medication recommendation framework by fusing molecular knowledge and electronic health records. Through a multimodal integration mechanism, HADRec jointly models clinical semantics encoded by LLaMA-7B and molecular features encoded by ChemBERTa. This enables cross-modality alignment between patient states and pharmacological mechanisms, substantially improving the model’s capacity to understand therapeutic effects. Furthermore, the framework employs a top-down hierarchical predictor coupled with a novel consistency loss, ensuring all recommendations are logically coherent and strictly comply with the ATC taxonomy. Comprehensive evaluations on the MIMIC-III dataset demonstrate that HADRec achieves state-of-the-art performance across key metrics (Jaccard, F1, PR-AUC).  Ablation studies confirm the vital contribution of the framework’s core components, which include molecular knowledge representation, multimodal fusion, hierarchical prediction architecture, and the consistency loss. Visualizations of the attention mechanism further reveal that the model learns an interpretable, “parent-to-child” reasoning pathway aligned with medical ontology. Threshold sensitivity analysis underscores the robustness of HADRec, which sustains high performance across a broad range of decision boundaries, with a simple validation-based threshold already achieving near-optimal results and effectively handling class imbalance. Counterfactual evaluation further provides strong evidence of causal and clinically aligned reasoning, showing that HADRec can disentangle disease-specific treatments from general care and adapt its recommendations in a consistent and interpretable manner. Multi-center validation and calibration analysis further demonstrate the model’s robustness and reliability under distribution shifts, indicating strong generalization and well-calibrated predictive confidence across clinical settings.

In summary, by grounding recommendations in molecular structures and enforced hierarchical logic, the framework establishes a robust paradigm for pharmacologically plausible and safer AI-driven prescriptions. Methodologically, it provides a transferable paradigm for embedding domain knowledge (e.g., taxonomies, ontologies) into neural architectures, paving the way for more reliable, interpretable, and knowledge-aware AI systems in healthcare.

\section*{CRediT authorship contribution statement}
Junke Wang: Writing – original draft, Visualization, Validation, Data curation.
Hongshun Ling: Writing – review \& editing, Methodology, Supervision, Conceptualization.
Li Zhang: Writing – review \& editing,Supervision, Conceptualization, Project administration.
Jinjing Wu: Writing – review \& editing.
Tong Shao: Writing – review \& editing.
Fang Wang: Supervision, Project administration.
Yuan Gao: Supervision, Project administration.

\section*{Declaration of competing interest}
The authors declare that they have no known competing financial interests or personal relationships that could have appeared to influence the work reported in this paper.

\section*{Acknowledgments}
This research is supported by Fundamental Research Funds for the Central Universities, South-Central Minzu University (No. CZQ24015), Key Research and Development Program of Hubei Province (No. 2022BAA037).

\section*{Data availability}
In this study, we used the MIMIC-III dataset from MIT’s Computational Physiology Laboratory. Code and data for this research are available at: https://github.com/wjk78623-cell/HADRec/tree/main

\appendix
\section{Longitudinal EHR Context Statistics}
\label{app1}
\begin{figure}[H]
\centering
\includegraphics[width=1\textwidth]{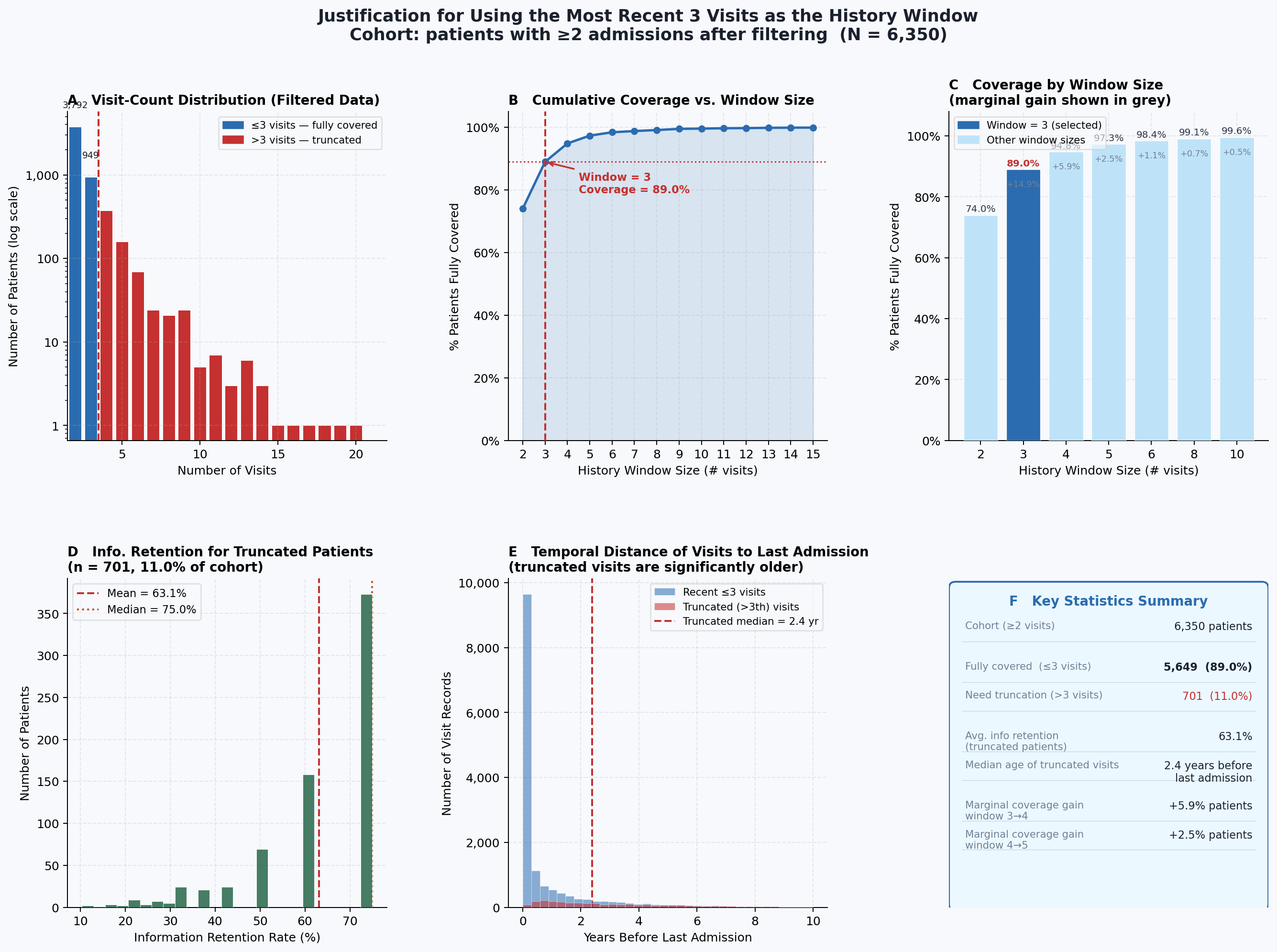}
\caption{Longitudinal EHR context analysis across different visit window configurations on the MIMIC-III cohort (N=6,350). (A) Distribution of patient counts by number of visits (log scale). (B) Cumulative patient coverage as a function of window size; window size 3 covers 89.0\% of patients. (C) Marginal coverage gain diminishes rapidly beyond window size 3. (D) Information retention rate for truncated patients (n=701) with mean 63.1\%. (E) Temporal distance of visits to last admission; truncated visits are significantly older (median 2.4 years). (F) Key summary statistics.}\label{fig5}
\end{figure}

\FloatBarrier     
\bibliographystyle{elsarticle-num-names}
\bibliography{cas-refs}

\end{document}